\documentclass{article}

\usepackage[final]{neurips_2020}
\pdfoutput=1
\usepackage[utf8]{inputenc} 
\usepackage[T1]{fontenc}    
\usepackage{hyperref}       
\usepackage{url}            
\usepackage{booktabs}       
\usepackage{amsfonts}       
\usepackage{nicefrac}       
\usepackage{microtype}      

\title{Robust Image Captioning}

\author{%
  Daniel Yarnell \\
  Department of Computer Science\\
  Central Michigan University\\
   \\
   \And
    Xian Wang \\
  Department of Computer Science\\
  Central Michigan University\\
   \\
}

\begin{document}

\maketitle
 \begin{abstract}
     Automated captioning of photos is a mission that incorporates the difficulties of photo analysis and text generation. One essential feature of captioning is the concept of attention: how to determine what to specify and in which sequence. In this study, we leverage the Object Relationusing adversarial robust cut algorithm, that grows upon this method by specifically embedding knowledgeabout the spatial association between input data through graph representation. Our experimental study represent the promising performance of our proposed method for image captioning. 
 \end{abstract}
\section{Introduction}
Formulating a description of an image is called image captioning. Image captioning needs identifying the significant objects, their properties, and their associations in an image. It also requires to develop syntactically and semantically precise expressions. Deep-learning-based methods are able to managing the difficulties and challenges of image captioning.There are two general paradigms in current image captioning methods: top-down and bottom-up. The top-down methods \cite{mao2014deep,karpathy2015deep} begins from an image and converts it into terms, while the bottom-up one \cite{farhadi2010every,kuznetsova2012collective} first formulates words explaining several components of an image and then merges them. Language models are utilized in both techniques to generate cohesive expressions. The state-of-the-art is the top-down framework that there is an end-to-end formulation from an image to a word in accordance with deep neural networks and all the aspects of the neural network can be acquired from training samples \cite{aneja2018convolutional}. We aim to better comprehend the importance of producing novel captions when we use the benchmark dataset MS COCO \cite{COCO} for automatic image captioning. In the past, several papers proposed composing image captions by finding similar images and copying their captions \cite{ordonez2011im2text,hodosh2013framing}. In large datasets like MS COCO \cite{COCO}, containing thousands of captions, the likelihood of finding a suitable caption increases, making these approaches more useful. Vinyals et al.~\cite{vinyals2014show} discovered that about $80\%$ of the image captions generated by their approach were the same as image captions in the MS COCO training dataset. This provides evidence of the effectiveness of copying captions. Although approaches based solely on copying the original captions may not perform as well as those that can additionally generate original captions if images in the MS COCO dataset contain too much diversity or capture many rare  Inspired by the success of leveraging structured prediction in different problems, we employ it as our criteria and propose a novel method for image captioning with memory and time efficiency.
\section{Related Works}
Recently, there has been significant attention to the automatic generation of captions
\cite{karpathy2014deep,lebret2015phrase,lebret2014simple,lazaridou2015combining}. Karpathy as a significant exceptionto this worldwide representation technique, retrieved characteristics from many image zones according to anR-CNN object detector and produced distinct captions for the zones. As a distinct caption was created for each zone, though, the spatial association between the discovered items was not modeled.
The proliferation of caption datasets has contributed to this surge in research \cite{ordonez2011im2text,young2014image,jianfu2015,COCO}, and new learning techniques \cite{Hinton,hochreiter1997long}. Recent proposals use many similar methods, including deep-learned image features \cite{Hinton,simonyan2014very}, and language models using maximum entropy \cite{fang2014captions}, recurrent neural networks \cite{chen2014learning,karpathy2014deep}, and LSTMs \cite{vinyals2014show,donahue2014long}.
Fang et al.in \cite{farhadi2010every} created image characterization by first tracking phrases related to various areas insidethe photo. The spatial association was carried out by employing a fully convolutional neural network to thepicture and producing spatial reaction maps for the goal phrases. Here again, the authors did notspecifically model any connections between the spatial areas.  
All of these methods are characterized by the ability to generate novel captions.
Several early neural simulations for image captioning encrypted visual knowledge utilizingone characteristic vector denoting the object in general, and therefore did not exploit knowledgeabout items and their spatial associations.

\section{Structured Prediction}
Structured prediction is increasingly vital for machine learning applications in computer vision \cite{kolmogorov2004energy}, natural language processing \cite{flake2004graph}, and other domains \cite{blum2001learning}. Many machine learning challenges include creating joint predictions over a set of mutually dependent output factors. The dependencies between output factors can be presented by a structure, such as a sequence, a graph, or a tree. Structured prediction methods have been developed for challenges of this type, and they have been demonstrated to be effective in several application domains, e .g natural language processing, bioinformatics, and computer vision. Graphical model methods and learning to search methods are two families of algorithms for these problems.
\cite{AAAI1817309} explored a robust method for learning to make min-cuts in graphs.
It works by making worst-case approximations to the training labels. 
This has advantages conceptually, supplying tight bounds on losses, and in practice, as demonstrated by our studies. ARC \cite{AAAI1817309,behpour2018arc,behpour2019active} models the multi-label classification as a complete graph. Inspired by this idea, we employ this approach for image captioning using convolutional Neural Networks (CNN).
\section{Experiments}
We performed studies on the MS COCO dataset \cite{lin2014microsoft}. Our train/validation/test splits follow \cite{xu2015show,karpathy2015deep}.We use 113287 training images, 5000 images for validation, and 5000 for testing. Thereafter, we will denote our method as CNN+ActiveARC, and the approach leveraging CNN with the attention as CNN+Attn. In table 1 and 2, Comparison of various approaches on standard evaluation metrics: BLEU-1 (B1), BLEU-2 (B2), BLEU-3 (B3),BLEU-4 (B4), METEOR (M), ROUGE (R), CIDEr (C) and SPICE (S). Our proposed method CNN+ARC attains comparablesuccess and outperforms other approaches. We begin with a CNN containing masked convolutions and fully connected layers. Then, we include more functions to our network like weight normalization, dropout,residual connections and attention incrementally. 
We investigated a mixture of convolutional neural network and graph model approach for image
captioning and showed that it outperforms other method. We explore active learning aspect in this direction following \cite{behpour2019active, behpour2019active1} for our future studies.
\begin{table}[h]
\centering
\label{tab:ex1}
 \caption{\small Comparison of different methods on standard evaluation metrics on MS-COCO dataset, Beam size=2}
\begin{tabular}{ | c | c | c | c | c | c | c | c | c |}
\hline
\tiny
Method & B1 & B2 & B3 & B4 & M & R & C & S \\
\hline\hline
LSTM &.715 &.545& .407 &.304 &.248& .526 &.940 &.178 \\
 CNN & .712 & .541 & .404 & .303 & .248 & .527 &.937 &.178 \\
  CNN+Attn & .718 &.549 &.411& .306& .248& .528& .942& .177 \\
CNN+ARC  &.731 &.612 &.471& .389& .314& .605& .953& .322 \\
 \hline

\end{tabular}
\end{table}

 \begin{table}[h]
\centering
\label{tab:ex2}
 \caption{\small Comparison of different methods on standard evaluation metrics on MS-COCO dataset, Beam size=4}
\begin{tabular}{ | c | c | c | c | c | c | c | c | c |}
\hline
\tiny
Method & B1 & B2 & B3 & B4 & M & R & C & S \\
\hline\hline
LSTM & 714 &.543 &.410& .311 &.250 &.529& .951 &.179 \\
 CNN & .706 & .533& .400& .302 &.247& .522& .925& .175 \\
  CNN+Attn & 718 & .550& .415 &.314 &.249& .528 &.951 &.179 \\
CNN+ARC  &.742&.609 &.466& .358& .309& .612& .932& .331 \\
 \hline

\end{tabular}
\end{table}
\bibliography{egbib3}
\bibliographystyle{apa-good}
\newpage

\end{document}